\documentclass{article}

\PassOptionsToPackage{square,sort,comma,numbers}{natbib}

\usepackage[final]{nips_2018}
\usepackage{times}
\usepackage{epsfig}
\usepackage{graphicx}
\usepackage{amsmath}
\usepackage{amssymb}
\usepackage{mathrsfs}
\usepackage{physics}
\usepackage{mathtools}

\usepackage[utf8]{inputenc} 
\usepackage[T1]{fontenc}    
\usepackage{hyperref}       
\usepackage{url}            
\usepackage{booktabs}       
\usepackage{amsfonts}       
\usepackage{nicefrac}       
\usepackage{microtype}      

\usepackage{todonotes}
\usepackage{multirow}
\usepackage{adjustbox}
\usepackage{tikz}

\title{Fourier-Domain Optimization for Image Processing}

\author{Majed El Helou\\
EPFL\\
Switzerland\\
{\tt\small majed.elhelou@epfl.ch}
\And
Frederike D{\"u}mbgen\\
EPFL\\
Switzerland\\
{\tt\small frederike.duembgen@epfl.ch}
\AND
Radhakrishna Achanta\\
Senior Scientist, SDSC\\
Switzerland\\
{\tt\small radhakrishna.achanta@datascience.ch}
\And
Sabine S{\"u}sstrunk\\
EPFL\\
Switzerland\\
{\tt\small sabine.susstrunk@epfl.ch}
}

\DeclareMathOperator{\circop}{circ}
\DeclareMathOperator{\fft}{fft2}
\DeclareMathOperator{\ifft}{ifft2}
\DeclareMathOperator{\psfotf}{psf2otf}

\newcommand{\matlab}{\texttt{MATLAB}}
\newcommand{\python}{\texttt{python}}

\newcommand{\fourier}{\mathcal{F}}

\newcommand{\inimage}{\mathbf{N}}
\newcommand{\outimage}{\mathbf{R}}

\newcommand{\argmin}{\arg\!\min}

\newcommand{\inR}[1]{\ensuremath{\in \mathbb{R}^{#1}}}

\begin{document}

\maketitle

\begin{abstract}
    Image optimization problems encompass many applications such as spectral fusion, deblurring, deconvolution, dehazing, matting, reflection removal and image interpolation, among others. With current image sizes in the order of megabytes, it is extremely expensive to run conventional algorithms such as gradient descent, making them unfavorable especially when closed-form solutions can be derived and computed efficiently. This paper explains in detail the framework for solving convex image optimization and deconvolution in the Fourier domain. We begin by explaining the mathematical background and motivating why the presented setups can be transformed and solved very efficiently in the Fourier domain. We also show how to practically use these solutions, by providing the corresponding implementations\footnote{{\matlab} and {\python} implementations: \url{https://github.com/duembgen/fourier-deconv}}. The explanations are aimed at a broad audience with minimal knowledge of convolution and image optimization.
    \textbf{The eager reader can jump to Section \ref{sec:optimization} for a footprint of how to solve and implement a sample optimization function, and Section \ref{sec:HQS} for the more complex cases.}
\end{abstract}

\section{Introduction}
A wide range of image processing applications are tackled in the literature by minimizing a carefully-defined loss function. These application-specific loss functions generally have one thing in common: the optimization variable is of the same order of magnitude in size as the image, and often it is the output image itself. This poses a big computational challenge when considering large images. Given that image sizes are consistently increasing, such optimization-based solutions are not suitable for deployment on low energy and mobile devices.

To remedy the computational problem, there exist different approaches that attempt to avoid explicitly formulating an optimization problem and settle instead for sub-optimal heuristics. This is because the current optimization schemes, which are essentially gradient descent based, are unable to converge fast enough with megapixel images. On the other hand, a high resolution is necessary for a multitude of computer vision applications as well as for better user experience. So an ideal solution would be capable of solving the formulated image optimization problem even for large image sizes. The bottom line is we desire high quality optimization without sacrificing computational efficiency or user experience. 

What we propose in this paper is not a novel method, rather a thorough explanation, visualization, and proof of validity of a very powerful optimization solver based on the use of the Fourier transform. In fact, the technique we explain has been central to several recent works of research~\cite{deconvHyperLap,IRCNN,l0smooth,antiGlare}. However, a thorough explanation of the method, including its mathematical background and implementation details, is lacking in the literature. Using a simple non-blind deconvolution problem as an example, we outline the details of the method, and elaborate upon the implementation details, in Section~\ref{sec:deconv}. After the theoretical and practical grounds are established, we present and explain all the steps involved in solving the image optimization problem in Section \ref{sec:optimization}. We also provide publicly accessible code to complement the paper.

The presented method addresses optimization problems that have a closed-form solution. In cases where there are inter-pixel relations that cannot be translated to convolutions, for example resulting from pixel-position-dependent mappings or complex cost functions, the direct method does not apply. For such situations we additionally address "half-quadratic splitting" techniques, which allow to split (as the name suggests) the optimization into two steps, one of which can typically be solved in closed form, and the other with a traditional gradient descent approach. The final solution for such problems is obtained by iterating over these two steps~\cite{nick, deconvHyperLap, kruse2017learning, IRCNN}. More details are provided in Section~\ref{sec:HQS}.


\section{Simple Non-blind Deconvolution} \label{sec:deconv}
In what follows, we assume that the reader has some background in basic linear algebra and image convolution-deconvolution. We assume that the number of pixels in an image $\mathbf{N}$ is $n=p \cdot q$. We treat one-channel images, and assume the pixel values to lie in $\mathbb{R}$.

\subsection{Mathematical Framework} \label{subsec:mathframe}

In the simple case of deconvolution, we observe an image $\outimage \inR{p'\times q'}$, which is the result of the convolution between an image $\inimage \inR{p\times q}$ and a convolution kernel $\mathbf{k}\inR{m\times n}$. The size of the observed image is reduced with respect to the input image such that $p'= p - (m-1)$ and $q' = q - (n-1)$. This size reduction is commonly avoided by zero-padding. The goal is to recover an accurate approximation of the input image $\inimage$. The kernel $\mathbf{k}$ can be the point spread function (PSF) of an acquisition system, omnipresent in coded aperture systems for instance. In non-blind deconvolution, we assume $\mathbf{k}$ to be known.

We start with an example convolution between a $1\times2$ kernel and a $3\times4$ matrix (our image), 
\begin{equation}
     \outimage = \mathbf{k} \circledast \inimage =
    \begin{bmatrix}
    -1 & 1
    \end{bmatrix}
    \circledast
    \begin{bmatrix}
    a_1 & b_1 & c_1 & d_1 \\
    a_2 & b_2 & c_2 & d_2 \\
    a_3 & b_3 & c_3 & d_3
    \end{bmatrix}
    =
    \begin{bmatrix}
    a_1-b_1 & b_1-c_1 & c_1-d_1 \\
    a_2-b_2 & b_2-c_2 & c_2-d_2 \\
    a_3-b_3 & b_3-c_3 & c_3-d_3
    \end{bmatrix},
\end{equation}\noindent
where $\circledast$ stands for 2D convolution.
This convolution can be written in terms of a matrix multiplication. For this, let us define the flipped and zero-filled kernel $\mathbf{k}' \inR{n}$ which in our example is given by:
\begin{equation}
    \mathbf{k}' = \begin{bmatrix}
    1, -1, 0, 0, 0, 0, 0, 0, 0, 0, 0, 0
    \end{bmatrix},
\end{equation}\noindent
where the mirroring is needed to mimic the convolution operation (mirroring and sliding), and the zero-padding is necessary to fit the image dimensions, as we will see shortly. The length of $\mathbf{k}'$ corresponds to the number of pixels in our image ($3 \cdot 4=12$).
We also need to reshape the matrix into a column vector in row-wise order, yielding:
\begin{equation}
    \inimage^{v} = \begin{bmatrix} a_1, b_1, c_1, d_1, a_2, b_2, c_2, d_2, a_3, b_3, c_3, d_3 \end{bmatrix}^T \inR{n}.
\end{equation}

We define the mapping $\circop:$ $\mathbb{R}^n \mapsto \mathbb{R}^{n\times n}$ which takes an input vector of length $n$ and maps it to a matrix of size $n\times n$ such that the matrix is circulant, and made up of integer shifts of the input vector. The operator applied to a 3 dimensional vector yields, for instance:
\begin{equation}
\circop(\begin{bmatrix} x_1 & x_2 & x_3 \end{bmatrix}) =
    \begin{bmatrix}
    x_1 & x_2 & x_3 \\
    x_3 & x_1 & x_2 \\
    x_2 & x_3 & x_1
    \end{bmatrix}.
\end{equation}

This allows us to rewrite the convolution equation in terms of a matrix multiplication:
\begin{equation} \label{eq:matrixMul}
    \outimage_{nv}^{v} = \circop(\mathbf{k}')  \inimage^{v},
\end{equation}
where $\outimage_{nv}^{v} \inR{n}$ is the vectorized result of the  \textit{nonvalid} convolution and $\circop(\mathbf{k}')$ is, in this particular example: 
\begin{equation}
    \circop(\mathbf{k}') = 
    \begin{bmatrix}
    1 & -1&  0& \cdots& 0 \\
    0& 1& -1& \ddots & \vdots \\
    \\
    \vdots& \ddots& \ddots& \ddots& 0 \\
    \\
    0& \cdots& 0& 1& -1 \\
    -1& 0& \cdots& 0& 1 
    \end{bmatrix}.
\end{equation}\noindent
The appellation \textit{nonvalid} is due to boundary effects: unlike $\outimage$ which is $p' \times q'$, $\outimage_{nv}$ is $p' \times q'+1$, with the extra column appended as an artifact of the convolution (dependent on the kernel size). For instance, in our original example, the result is:
\begin{equation}
    \outimage_{nv} = 
    \begin{bmatrix}
    a_1-b_1 & b_1-c_1 & c_1-d_1 & d_1-a_2 \\
    a_2-b_2 & b_2-c_2 & c_2-d_2 & d_2-a_3 \\
    a_3-b_3 & b_3-c_3 & c_3-d_3 & d_3-a_1
    \end{bmatrix}.
\end{equation}
Notice, however, that $\outimage$ is exactly equal to the first $n-1$ columns of $\outimage_{nv}$. This is what is referred to as the "circular boundary conditions" when using this framework in the literature \cite{deconvHyperLap}.

We could solve Eq. \eqref{eq:matrixMul} for $\inimage^v$ by inverting the circulant matrix. Note however that the circulant matrix is of size $\mathbb{R}^{n\times n}$, and such an inversion is very costly. Therefore, we move to the Fourier domain, exploiting the following well-known theorem. 
\\ 

\textbf{Theorem 1} The discrete Fourier transform (DFT) matrix diagonalizes any circulant matrix. In other words, any matrix in the frequency domain, which is the transform of a circulant matrix, is a diagonal matrix. \\

Based on Theorem 1, $\circop(\mathbf{k}')$ in Eq. \eqref{eq:matrixMul} becomes a diagonal matrix in the Fourier domain (denoted by $\fourier$). 
The multiplication in the time/space domain naturally becomes a convolution in the Fourier domain:
\begin{equation} \label{eq:matrixFconv}
    \fourier(\outimage_{nv}^{v}) = \fourier(\circop(\mathbf{k}')) \circledast \fourier(\inimage^{v}).
\end{equation}

Two important things to note here are that, first, we are consistently utilizing the DFT across Eq. \eqref{eq:matrixFconv}. And second, the convolution is between a diagonal matrix and a column vector, making it effectively nothing more than a point-wise multiplication of their entries. This point-wise multiplication is seen in Eq. \eqref{eq:psf2otf} but where the non-trivial entries are compactly collected into two small matrices instead of one large diagonal matrix and a corresponding-length vector. This means that every entry in $\fourier(\inimage^{v})$ is multiplied by one diagonal element in $\fourier(\circop(\mathbf{k}'))$ to obtain the corresponding element in $\fourier(\outimage_{nv}^{v})$. Therefore, a simple point-wise division is enough to recover $\fourier(\inimage^{v})$, and applying the inverse Fourier transform yields $\inimage^{v}$, that is, the deconvolved image in vectorized form.

\subsection{Kernels with Two Dimensions} \label{subsec:2Dkernel}
The generalization to 2D convolution kernels is straight-forward. The only difference relative to the previous section is in the construction of $\mathbf{k}'$. Taking as an example a 2D kernel $\mathbf{k}$:
\begin{equation}
    \mathbf{k} = \begin{bmatrix}
    -1 & -2 \\
    -3 & -4
    \end{bmatrix},
\end{equation}\noindent
the corresponding $\mathbf{k}'$, for the same image size of $3\times4$, is given by:
\begin{equation}
    \mathbf{k}' = \begin{bmatrix}
    -4, -3, 0, 0, -2, -1, 0, 0, 0, 0, 0, 0
    \end{bmatrix}.
\end{equation}\noindent
In general, $\mathbf{k}'$ always contains the unrolled version of a doubly-mirrored or flipped version of $\mathbf{k}$ called $\mathbf{k_f}$:
\begin{equation}
    \mathbf{k_f} = \begin{bmatrix}
    -4 & -3 \\
    -2 & -1
    \end{bmatrix}.
\end{equation}\noindent
The way $\mathbf{k_f}$ is unrolled is row by row: the first row is filled into the first entries of $\mathbf{k}'$, then the second row is filled in but beginning at entry $q+1$, where $q$ is the number of columns in the image (i.e. the size of an image row). In this example, $-2$ is filled in at position $4+1=5$. This procedure is repeated until all the rows of $\mathbf{k_f}$ have been traversed and filled in $\mathbf{k}'$. The remainder of $\mathbf{k}'$ is left as the original zero filling. 

For completeness, it is theoretically possible, although practically not relevant, to have a convolution kernel wider than the image and thus not fitting in this construct. In that case, however, there is not a single valid convolution being carried out, since the size of the valid output normally is $p' \times q'$ where $p'= p - (m-1)$ and $q' = q - (n-1)$. Thus the valid convolution is the empty set whenever $m \geq (p+1)$ or $n \geq (q+1)$.

\subsection{Practical Implementation} \label{subsec:practicalImplementation}
In the previous section we have leveraged two properties of the Fourier transform: a time/space-domain convolution corresponds to a multiplication in the Fourier domain, and the DFT diagonalizes circulant matrices. However, an obvious issue in Eq. \eqref{eq:matrixMul} is the data complexity. For an image containing $n$ (possibly millions of) pixels, the matrix $\circop(\mathbf{k}')$ is of size $n\times n$, making it very expensive to take its Fourier transform or even to simply store it. However, once in the Fourier domain, only $n$ diagonal entries are non-zero, so the matrix is  sparse. In both {\matlab} and {\python} there exist libraries that can handle relatively large sparse matrices and can compute their inverses, however, these are (as of this date) limited to too small image sizes. In practical implementations, it is possible to bypass these steps and to go directly to the Fourier domain, which is what we explain in this section.

So instead of converting our images into vectorized forms, we can keep them in their original matrix form, and compute and arrange the diagonal entries of $\fourier(\circop(\mathbf{k}'))$ into a single matrix of same size as the images. 
It is also possible to work with the two-dimensional Fourier transform of the images, if the corresponding entries in the diagonal matrix ($\fourier(\circop(\mathbf{k}'))$ Eq. \eqref{eq:matrixFconv}) are adjusted. And by adjusted, we mean that a one-dimensional Fourier transform needs to be applied to the matrix-rearranged diagonal entries. This is exactly what $\psfotf$ does, by first distributing the entries of the input kernel $\mathbf{k}$ over a matrix of same size as the image, then applying the two-dimensional Fourier transform to that resulting matrix.
In the provided implementations, the convolution operation corresponds to:
\begin{equation} \label{eq:psf2otf}
    \fft(\outimage_{nv}) = \psfotf(\mathbf{k}_f, S=[p,q]) .* \fft(\inimage),
\end{equation}
where $\fft$ and $\psfotf$ are built-in functions, $.*$ is point-wise multiplication, $S=[p,q]$ holds the dimensions of the image, and $\mathbf{k}_f$ is the mirrored version of $\mathbf{k}$. All what $\psfotf$ does is compute the diagonal entries of the diagonal matrix $\fourier(\circop(\mathbf{k}'))$ by reshaping $\mathbf{k}$ into a $p\times q$ matrix and taking the DFT of that matrix. The deconvolution simply becomes:
\begin{equation}
    \fft(\inimage) = \fft(\outimage_{nv})./\psfotf(\mathbf{k}_f, S),
\end{equation}
where the division is again point-wise. The final deconvolved image can be obtained by going back to the spatial domain using the inverse Fourier transform ($\ifft$).

\begin{figure}
\centering 
\begin{minipage}{.47\linewidth}
\includegraphics[width=\linewidth]{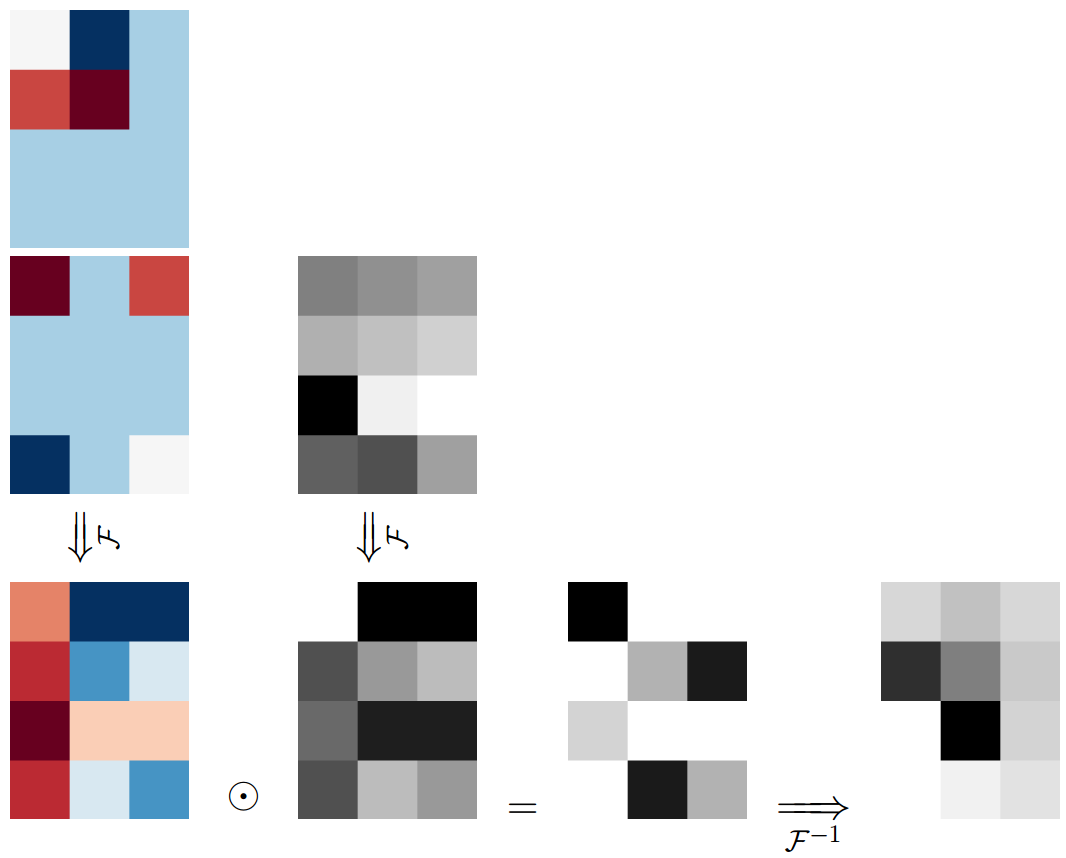} 

(a) Visualization of the operation of $\psfotf$. In colors are from top to bottom: the convolution kernel, the same kernel with entries rearranged, the two-dimensional Fourier transform of the latter matrix. In grey are similarly the input image, and its two-dimensional Fourier transform. 
\end{minipage}\hfill
\begin{minipage}{.45\linewidth}
\includegraphics[width=\linewidth]{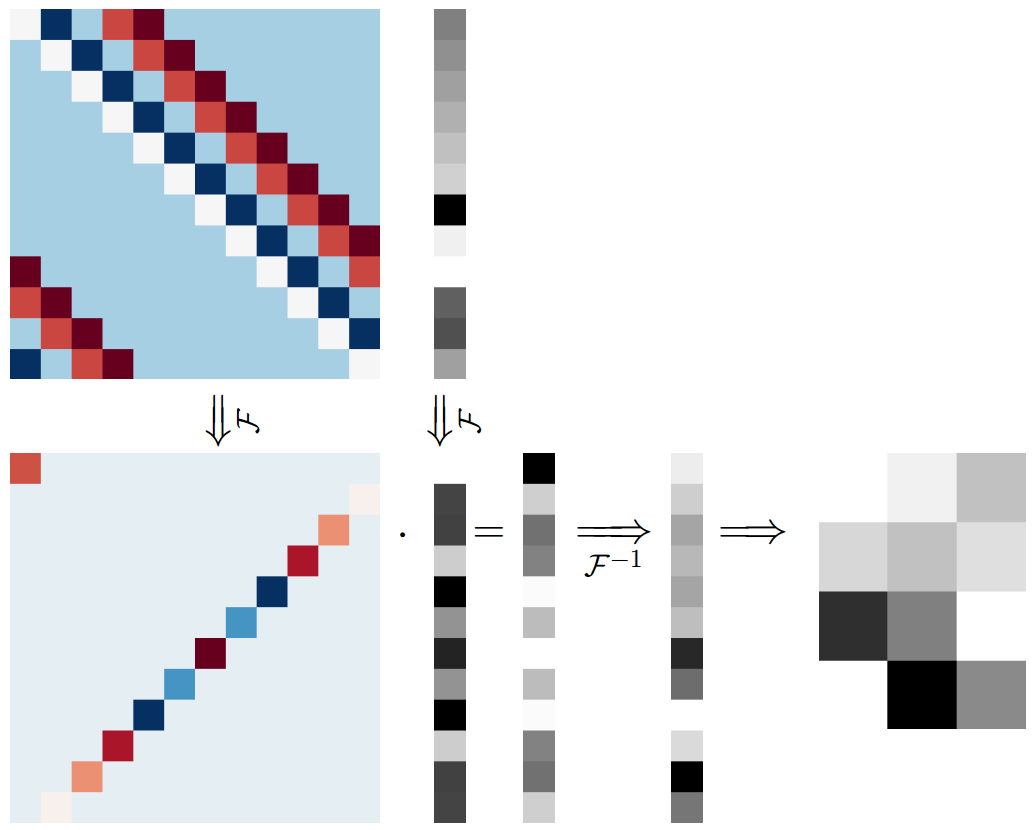} 

(b) Visualization of the large convolution in the Fourier domain between the Fourier transform of the circulant matrix and the Fourier transform of the vectorized image. This effectively amounts to a point-wise multiplication.
\end{minipage} 
\caption{Visualizations of $\python$ implementations of both (a) the $\psfotf$ approach and (b) the large matrix multiplication on vectorized images.}
\label{fig:psf2otf}
\end{figure}

\subsection{Remarks}
1) Note that the one-dimensional kernel $\mathbf{k}$ we used for illustration implements an image gradient in the horizontal direction (vertical edge detector). This will be used in the next section.

2) Circulant matrices have some well-defined properties, which are exploited in the proposed solution in Section \ref{sec:optimization}. These properties are outlined in the theorems below. \\ 

\textbf{Theorem 2} The transpose $\mathbf{A}^T$ of a circulant matrix $\mathbf{A}=\circop(\mathbf{a})$ is also a circulant matrix.\\
\textbf{Proof} The proof is trivial and this can easily be visualized. Let $\mathbf{y}$ be the bottom row of $\mathbf{A}$, and $\mathbf{y}'$ its mirrored version, we then have $\mathbf{A}^T = \circop(\mathbf{y}')$, and thus $\mathbf{A}^T$ is circulant. \\

\textbf{Theorem 3} Multiplying a circulant matrix $\mathbf{A}=\circop(\mathbf{a})$ by a circulant matrix $\mathbf{B}$ yields a circulant matrix $\mathbf{A}\cdot\mathbf{B}$. \\
\textbf{Proof} We know from Theorem 2 that $\mathbf{B}^T$ is also circulant; $\mathbf{B}^T = \circop(\mathbf{b})$. (a) This implies that not only the rows but also the columns of $\mathbf{B}$ are circulant variants of a single vector. Let $\mathbf{a}(-i)$ denote the circular shifting of the vector $\mathbf{a}$ by $i$ steps. (b) The inner product $\mathbf{a}(-i)\cdot\mathbf{b}(-j)$ only depends on the difference $(i-j)$. (a) \& (b) imply by construction that $\mathbf{A}\cdot\mathbf{B} = \circop(\mathbf{c})$ where $\mathbf{c} = 
\begin{bmatrix} \mathbf{a} \cdot \mathbf{b} & \mathbf{a} \cdot \mathbf{b} (-1) & \dots & \mathbf{a} \cdot \mathbf{b} (-(N-1)) 
\end{bmatrix}$ and $N$ is the number of columns of $\mathbf{B}$.\\

\textbf{Theorem 4} The sum of a circulant matrix $\mathbf{A}=\circop(\mathbf{a})$ and a circulant matrix $\mathbf{B}=\circop(\mathbf{b})$ is also a circulant matrix. And $circ$ is a linear function. \\
\textbf{Proof} The sum $\mathbf{A}+\mathbf{B}$ is trivially the circulant matrix $\circop(\mathbf{a}+\mathbf{b})$. And also trivially $\circop(\lambda\mathbf{a}) = \lambda\circop(\mathbf{a})$ for any constant $\lambda$. \\

\section{Convex Optimization Solution} \label{sec:optimization}

\subsection{Mathematical Framework}
In this section, we explain how closed-form solutions to convex image optimization problems can be computed efficiently ($\mathcal{O}(n\log{}n)$), instead of running a  descent algorithm. Such approaches have been used in \cite{deconvHyperLap}, \cite{l0smooth}, and \cite{antiGlare}, however, a clear explanation and implementation details are missing in the literature. \\

A common denominator of many image processing problems is that a solution can be found by minimizing a cost function, tailored to the problem, over a high-dimensional argument, typically the image itself. One example of such cost functions is:
\begin{equation} \label{eq:lossFunc}
    \mathcal{L}(\mathbf{N}) = \lambda || \outimage - \mathbf{b}\circledast \mathbf{N} ||_F^2 + || \mathbf{X} - \nabla_x \mathbf{N} ||_F^2,
\end{equation}
\noindent where the first and second cost terms enforce data consistency between the given guiding matrices $\outimage$ and $\mathbf{X}$, and the image on one side and the image
gradient on the other side, respectively, weighted by the regularization factor $\lambda \in \mathbb{R}$.
This cost function can be augmented with any number of cost terms including additional kernels and guiding matrices. Remember that the gradient $\nabla_x$ is also a convolution with the kernel discussed in Section \ref{sec:deconv}. Note that we cannot allow point-wise operations on the optimization argument if we want a closed-form solution, which is discussed in the following section, to exist. Otherwise, half-quadratic splitting is required for the optimization solution to benefit from the proposed approach. 

The loss function in Eq.~\eqref{eq:lossFunc} is a summation of squared Frobenius norms and is thus convex with a unique solution that can be found in closed form. The optimal solution can be found at the zero-gradient location, so we differentiate the loss function with respect to our optimization variable image $\mathbf{N}$. To do this, we write the loss in terms of matrix multiplications (with no convolution operations) and vectorized images. Based on our analysis in the previous section, this leads to the following, equivalent loss:
\begin{equation}
    \mathcal{L}(\inimage) = \mathcal{L}(\inimage^{v}) = \lambda || \outimage^{v} - \mathbf{B}  \inimage^{v} ||_F^2 + || \mathbf{X}^{v} - \mathbf{K}  \inimage^{v} ||_F^2,
\end{equation}\noindent
where $\mathbf{B}=\circop(\mathbf{b}')$ and $\mathbf{K}=\circop(\mathbf{k}')$ are defined as in Section \ref{sec:deconv}. Note that they do not have to be computed in practice, as we can directly use their Fourier representation, as explained in Section \ref{subsec:practicalImplementation}. Taking the partial derivative of our loss function, we get:
\begin{equation}
    \pdv{\mathcal{L}}{\inimage^{v}} = 2\lambda \mathbf{B}^T(\mathbf{B} \inimage^{v}-\outimage^{v})  +  2\mathbf{K}^T(\mathbf{K} \inimage^{v}-\mathbf{X}^{v}).
\end{equation}

Expanding and setting the derivative to zero yields:
\begin{equation}
    2\lambda \mathbf{B}^T\mathbf{B}\inimage^{v} - 2\lambda \mathbf{B}^T \outimage^{v}  +  2\mathbf{K}^T\mathbf{K}\inimage^{v} - 2\mathbf{K}^T\mathbf{X}^{v} = 0,
\end{equation}
\noindent
which can be rearranged in the standard form that is analyzed in Section \ref{sec:deconv}:
\begin{equation}\label{eq:image-opt}
    \mathbf{C} \inimage^{v} = \widetilde{\outimage}^{v}
\end{equation}\noindent
where $\mathbf{C} = \lambda \mathbf{B}^T  \mathbf{B} + \mathbf{K}^T  \mathbf{K}$ is a circulant matrix based on Theorems 2, 3 and 4, and $\widetilde{\outimage}^{v} = \lambda \mathbf{B}^T \outimage^{v}  + \mathbf{K}^T \mathbf{X}^{v}$ is a known vector. 

\subsection{Practical Implementation}

We can solve Eq. \eqref{eq:image-opt} in the Fourier domain as derived in Section \ref{subsec:practicalImplementation}.
The Fourier domain equivalent of $\mathbf{C}$ is given by:
\begin{equation}
    \fourier(\mathbf{C}) = \lambda \fourier(\mathbf{B})^*  \fourier(\mathbf{B}) + \fourier(\mathbf{K})^*  \fourier(\mathbf{K}),
\end{equation}
where $^*$ denotes the conjugate or Hermitian transpose of a matrix. As before, we obtain a non-sparse representation of the Fourier coefficients by leveraging $\psfotf$:
\begin{equation}
    \fourier(\mathbf{B}) = \psfotf(\mathbf{b},S), \quad
    \fourier(\mathbf{K}) = \psfotf(\mathbf{k},S), \quad
\end{equation}
where $S$ is the size of the (non-vectorized) input image.
To sum up with, the optimal solution becomes:
\begin{equation}
    \boxed{
    \inimage_{opt} = \ifft \left( 
    \frac{\lambda \psfotf(\mathbf{b}_f,S)' .* \fft(\outimage)  +  \psfotf(\mathbf{k}_f,S)'.* \fft(\mathbf{X})}
    {( \lambda \psfotf(\mathbf{b}_f,S)' .* \psfotf(\mathbf{b}_f,S)  +  \psfotf(\mathbf{k}_f,S)' .* \psfotf(\mathbf{k}_f,S) )} \right),
    }
    \label{eq:fullSolution}
\end{equation}
\noindent
where the division and all multiplications are point-wise, and $'$ is the MATLAB symbol for the conjugate or Hermitian transpose. 

\section{GitHub demos} \label{sec:github}

On the GitHub page are made available a {\matlab} and a {\python} demo that can be downloaded and run without any additional setup or toolboxes needed. The demos are split into two parts A and B as follows. 

Part A illustrates the simple convolution operation described in Section \ref{sec:deconv}. The convolution is carried with the built-in functions, but also with the large matrix multiplication as well as in the Fourier domain. One can visualize the results, which are automatically printed, and notice that the nonvalid entries (last column in the chosen example) are different, but all valid entries are consistent. 

Part B illustrates how to solve a guided deblurring image optimization in the Fourier domain. The script reads a guide RGB image, and a blurry near-infrared (NIR) image. The deblurred output is the image minimizing the loss function $\mathcal{L}(\cdot)$ that is given by:
\begin{equation}
    \mathcal{L}(\mathbf{N}) = \lambda ||\mathbf{N_b}-\mathbf{b}\circledast \mathbf{N}||_F^2 + \sum_{i=x,y}||\nabla_i \mathbf{Y} - \nabla_i \mathbf{N}  ||_F^2,
\end{equation}
\noindent
where $\lambda$ is a regularization factor (set to 1), $\mathbf{N_b}$ is the blurry NIR input, and $\mathbf{b}$ the estimated blur. For simplicity, it is assumed to be Gaussian and constant across the entire image. $\mathbf{Y}$ is the luminance of the RGB guide and $\mathbf{N}$ is the optimization variable. The interested reader can refer to the following works for the more advanced and better performing versions of this multispectral deblurring loss function \cite{ICIP, zahra}.

\section{Half-Quadratic Splitting} \label{sec:HQS}
The previous sections explain how to write-out the closed-form solution of the optimization, and how to solve for it efficiently in the Fourier domain. However, as we discuss in our introduction, not any optimization can be solved directly with this approach, without half-quadratic splitting. Let us consider a general loss function we aim to minimize:
\begin{equation}
    \mathcal{L}(\mathbf{N}) = f_1( \mathbf{N} ) + f_2( \mathbf{N} ) ,
\end{equation}
where $f_2(\cdot)$ is non-quadratic in $\mathbf{N}$ and a solution cannot be written-out in closed form. This is where half-quadratic splitting comes into play. The loss $\mathcal{L}(\cdot)$ can be minimized iteratively by optimizing for $f_1(\cdot)$ and $f_2(\cdot)$ one after the other. For that, the optimization loss is rewritten in terms of $\mathbf{N}$ and $\mathbf{Z}$, where $\mathbf{Z}$ is a latent variable used to split the two functions in $\mathcal{L}(\cdot)$ for the iterative optimization. We obtain the new loss, which we minimize with respect to both $\mathbf{N}$ and $\mathbf{Z}$:
\begin{equation}
    \mathcal{L}_2(\mathbf{N},\mathbf{Z}) = f_1( \mathbf{N} ) + f_2( \mathbf{Z} ) .
\end{equation}
With this new definition, $\mathcal{L}_2(\cdot)$ can be minimized by minimizing over $\mathbf{N}$, which can be done efficiently using the approach we described in previous sections, since $f_2(\cdot)$ is independent of $\mathbf{N}$, then minimizing over $\mathbf{Z}$, for instance with descent approaches. However, this method fails to converge to the same solution as $\mathcal{L}(\cdot)$, since $\mathbf{N}$ and $\mathbf{Z}$ are not constrained to being equal. To enforce this and converge to the minimizer of $\mathcal{L}(\cdot)$, we add another term to the loss:
\begin{equation}
    \mathcal{L}_3(\mathbf{N},\mathbf{Z}) = f_1( \mathbf{N} ) + f_2( \mathbf{Z} ) + \beta ||\mathbf{N} - \mathbf{Z}||_F^2 .
\end{equation}
where $\beta$ is a small positive constant that is increased with every minimization iteration. As $\beta \to +\infty$, we obtain $\mathbf{N}=\mathbf{Z}$ (as a result of minimizing $\mathcal{L}_3(\cdot)$, since otherwise their difference makes the loss infinitely large) thus making $\mathcal{L}(\cdot)$ and $\mathcal{L}_3(\cdot)$ equivalent.
The solution is obtained by alternately minimizing over $\mathbf{N}$ and $\mathbf{Z}$ with the following steps (shown for time step $t$):
\begin{equation} \label{eq:step1}
    \mathbf{N}^t = \argmin_\mathbf{N} \mathcal{L}_3(\mathbf{N},\mathbf{Z}^{t-1}) = \argmin_\mathbf{N} \left( f_1( \mathbf{N} ) + \beta ||\mathbf{N} - \mathbf{Z}^{t-1}||_F^2 \right)
\end{equation}
\begin{equation} \label{eq:step2}
    \mathbf{Z}^t = \argmin_\mathbf{Z} \mathcal{L}_3(\mathbf{N}^{t},\mathbf{Z}) = \argmin_\mathbf{Z} \left( f_2( \mathbf{Z} ) + \beta ||\mathbf{N}^{t} - \mathbf{Z}||_F^2 \right)
\end{equation}
\begin{equation}
    \beta^t = g(\beta^{t-1})
\end{equation}    
where $g(\cdot)$ is an increasing function, and $\mathbf{Z}^{t-1}$ can be initialized randomly or with an educated guess when possible. Eq. \eqref{eq:step1} can be solved with the method presented in this paper, and Eq. \eqref{eq:step2} needs to be solved with descent algorithms.

The half-quadratic splitting approach we present belongs to the family of variable splitting optimization techniques, with the particularity that the splitting approach is designed specifically to leverage the closed-form Fourier solution. More precisely, the variable splitting optimization is carried with the penalty approach, where the penalty term added in $\mathcal{L}_3$ is responsible for constraining $N$ to be equal to $Z$. This constraint is made less and less loose as we increase the value of $\beta$ with every iteration. This behavior can be controlled by tuning the parameter $\beta$.

Another variable splitting strategy is similar to optimizing $\mathcal{L}_2$ under the constraint $N=Z$. The constrained optimization can be carried out with the augmented Lagrangian method, or ADMM (first appearing in \cite{ADMM}) when the constraint has more than one component vector and it is possible to alternate the optimization. For the purpose of this paper, we only present the half-quadratic splitting technique because its formulation allows the leveraging of the closed-form Fourier-domain solution that we present.

\section{Final Remarks}
\begin{enumerate}
    \item It is preferable, when a closed-form solution is readily available, to make use of it and avoid gradient descent approaches. The latter are far less efficient, and also require hyper-parameter tweaking. Obtaining the closed-form solution in the Fourier domain is itself more efficient both computationally ($\log{}(n)/n^2$ improvement) and memory-wise ($\mathcal{O}(n)$ instead of $\mathcal{O}(n^2)$) compared to matrix inversion.
    \item When a closed-form solution cannot be formulated, half-quadratic splitting can be used to divide the optimization into two steps, one having a closed-form solution and the other requiring traditional descent-based optimization.
    \item In some (simplistic) scenarios where the optimization has a "null-space", the optimal solution is non-unique. This is simpler to reason about with an example. One such case is when the optimization variable is always convolved in the loss function with a gradient kernel (ex: [-1 1]) that is symmetric in absolute value but of opposite signs on each side of the center. Such an operator is not affected by image-wide constant shifts, since they get added then removed by the convolution with [-1 1]. This means that any constant shift added to the optimal solution does not affect the loss value, and will itself also be an optimal solution too. This issue is a well-known problem in image deconvolution.
\end{enumerate}

\section{Conclusion}
In this paper, we present the framework for efficiently solving image deconvolution and, more generally, a variety of image optimization problems, in the Fourier domain. We explain the mathematical background in simple terms, as well as the steps involved in the derivation of the optimization solution. We provide an open-source {\matlab} and {\python} demo showing the presented theory in practice for two selected examples. Researchers can leverage this framework for solving other image-related problems, by formulating the matching optimization and solving it efficiently for very large image sizes. We also explain briefly the approach of half-quadratic splitting, which makes it possible to leverage the proposed method in the case of more general loss functions.

We invite the readers who implement this approach in other programming languages to contribute a sample implementation or demo similar to ours to the GitHub repository.

\section*{Acknowledgements}
The authors would like to thank Dr. Zahra Sadeghipoor and Dr. Nikolaos Arvanitopoulos for valuable discussions and advice.

{\small


\begin{thebibliography}{1}\itemsep=-1pt

\bibitem{nick}
N.~Arvanitopoulos, R.~Achanta, and S.~Susstrunk.
\newblock Single image reflection suppression.
\newblock In {\em Proceedings of the IEEE Conference on Computer Vision and
  Pattern Recognition (CVPR)}, pages 4498--4506, 2017.

\bibitem{ICIP}
M.~El~Helou, Z.~Sadeghipoor, and S.~S{\"u}sstrunk.
\newblock Correlation-based deblurring leveraging multispectral chromatic
  aberration in color and near-infrared joint acquisition.
\newblock In {\em IEEE International Conference on Image Processing (ICIP)},
  2017.

\bibitem{ADMM}
R.~Glowinski and A.~Marroco.
\newblock Sur l'approximation, par {\'e}l{\'e}ments finis d'ordre un, et la
  r{\'e}solution, par p{\'e}nalisation-dualit{\'e} d'une classe de
  probl{\`e}mes de dirichlet non lin{\'e}aires.
\newblock {\em Revue fran{\c{c}}aise d'automatique, informatique, recherche
  op{\'e}rationnelle. Analyse num{\'e}rique}, 9(R2):41--76, 1975.

\bibitem{deconvHyperLap}
D.~Krishnan and R.~Fergus.
\newblock Fast image deconvolution using hyper-laplacian priors.
\newblock In {\em Advances in Neural Information Processing Systems (NIPS)},
  pages 1033--1041, 2009.

\bibitem{kruse2017learning}
J.~Kruse, C.~Rother, and U.~Schmidt.
\newblock Learning to push the limits of efficient fft-based image
  deconvolution.
\newblock In {\em Proceedings of the IEEE Conference on Computer Vision and
  Pattern Recognition (CVPR)}, pages 4586--4594, 2017.

\bibitem{l0smooth}
Y.~Li and M.~S. Brown.
\newblock Single image layer separation using relative smoothness.
\newblock In {\em Proceedings of the IEEE Conference on Computer Vision and
  Pattern Recognition (CVPR)}, pages 2752--2759, 2014.

\bibitem{zahra}
Z.~Sadeghipoor~Kermani, Y.~Lu, E.~Mendez, and S.~S{\"u}sstrunk.
\newblock Multiscale guided deblurring: Chromatic aberration correction in
  color and near-infrared imaging.
\newblock In {\em European Conference on Signal Processing (Eusipco)}, 2015.

\bibitem{antiGlare}
T.~Sandhan and J.~Y. Choi.
\newblock Anti-glare: Tightly constrained optimization for eyeglass reflection
  removal.
\newblock In {\em Proceedings of the IEEE Conference on Computer Vision and
  Pattern Recognition (CVPR)}, pages 1241--1250, 2017.

\bibitem{IRCNN}
K.~Zhang, W.~Zuo, S.~Gu, and L.~Zhang.
\newblock Learning deep cnn denoiser prior for image restoration.
\newblock In {\em IEEE Conference on Computer Vision and Pattern Recognition
  (CVPR)}, pages 2808--2817, 2017.

\end{thebibliography}
}

\end{document}